\documentclass{ecai} 

\usepackage{latexsym}
\usepackage{amssymb}
\usepackage{amsmath}
\usepackage{amsthm}
\usepackage{booktabs}
\usepackage{enumitem}
\usepackage{graphicx}
\usepackage{color}
\usepackage[table,xcdraw]{xcolor}
\usepackage[table]{xcolor}
\usepackage{colortbl}
\usepackage{multirow}

\newcommand{\BibTeX}{B\kern-.05em{\sc i\kern-.025em b}\kern-.08em\TeX}

\begin{document}


\begin{frontmatter}




\title{Thinning a medical image segmentation model via dual-level multiscale fusion}


\author[A,B]{\fnms{Chengqi}~\snm{Dong}}
\author[A,B]{\fnms{Fenghe}~\snm{Tang}} 
\author[A,B]{\fnms{Rongge}~\snm{Mao}} 
\author[A,B]{\fnms{Xinpei}~\snm{Gao}} 
\author[A,B,C,D,E]{\fnms{S.Kevin}~\snm{Zhou}\orcid{0000-0002-6881-4444}\thanks{Corresponding Author. Email: skevinzhou@ustc.edu.cn.}\footnotemark}

\address[A]{School of Biomedical Engineering, Division of Life Sciences and Medicine, University of Science and Technology of China (USTC), Hefei, 230026, China}
\address[B]{Center for Medical Imaging, Robotics, Analytic Computing \& Learning (MIRACLE), Suzhou Institute for Advance Research, USTC, Suzhou, 215123, China}
\address[C]{Key Laboratory of Intelligent Information Processing of Chinese Academy of Sciences (CAS), Institute of Computing Technology, CAS, Beijing, 100190, China}
\address[D]{Jiangsu Provincial Key Laboratory of Multimodal Digital Twin Technology, Suzhou, 215123, China}
\address[E]{State Key Laboratory of Precision \& Intelligent Chemistry, USTC, Hefei, China}


\begin{abstract}

Medical image segmentation plays a pivotal role in disease diagnosis and treatment planning, particularly in resource-constrained clinical settings where lightweight and generalizable models are urgently needed. However, existing lightweight models often compromise performance for efficiency and rarely adopt computationally expensive attention mechanisms, severely restricting their global contextual perception capabilities. Additionally, current architectures neglect the channel redundancy issue under the same convolutional kernels in medical imaging, which hinders effective feature extraction. To address these challenges, we propose LGMSNet, a novel lightweight framework based on local and global dual multiscale that achieves state-of-the-art performance with minimal computational overhead. 
LGMSNet employs heterogeneous intra-layer kernels to extract local high-frequency information while mitigating channel redundancy. In addition, the model integrates sparse transformer-convolutional hybrid branches to capture low-frequency global information.
Extensive experiments across six public datasets demonstrate LGMSNet's superiority over existing state-of-the-art methods. In particular, LGMSNet maintains exceptional performance in zero-shot generalization tests on four unseen datasets, underscoring its potential for real-world deployment in resource-limited medical scenarios. The whole project code is in https://github.com/cq-dong/LGMSNet.
\end{abstract}

\end{frontmatter}


\section{Introduction}
Medical image segmentation enables clinicians to obtain objective references for regions of interest~\cite{tang2024cmunext,xing2024segmamba}, which is key to the early detection of malignant tumors or abnormal cells, which helps physicians in diagnosis and treatment~\cite{coates2015tailoring}. Although heavyweight models such as U-Net variants~\cite{ronneberger2015u,chen2021transunet} and fused Vision Transformers (ViTs)~\cite{dosovitskiy2020image} achieve high precision, their computational complexity and resource demands hinder equitable healthcare access. Although lightweight alternatives offer promise for mobile and edge deployment, they typically sacrifice performance or generalizability. This trade-off requires efficient, adaptive segmentation frameworks that balance accuracy, efficiency, and cross-domain robustness~\cite{chen2024tinyu}.
\begin{figure*}[htbp]
\centering
\includegraphics[width=0.95\textwidth]{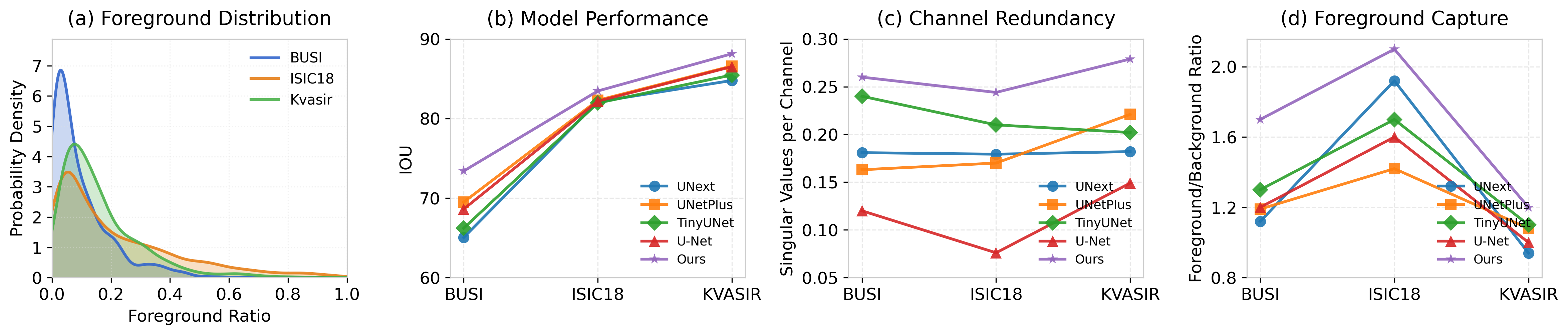}
\caption{Comparative analysis of medical image segmentation models on three modalities (BUSI, ISIC18, and Kvasir). (a) Probability density distribution of the lesion foreground scale ratio of three datasets. (b) Segmentation performance measured by IoU. (c) Channel redundancy is analyzed by the average singular value of each channel, where high values represent low redundancy. (d) Foreground capture capability is analyzed by the foreground-background eigenvalue ratio. Our method demonstrates excellent performance, utilizing channel redundancy information and stronger lesion capture.} 
\label{fig:svd}
\end{figure*}

In the classical design paradigm of medical segmentation, the U-Net~\cite{ronneberger2015u} adopts an encoder-decoder architecture, which has inspired the emergence of numerous convolutional networks~\cite{zhou2019unet++, oktay2018attention, isensee2021nnu}. However, these models achieve excellent segmentation performance at the cost of a large number of parameters and computational resources. More lightweight approaches typically reduce the number of parameters through network width compression and depthwise convolutions~\cite{valanarasu2022unext, tang2024cmunext, chen2024tinyu}. Nevertheless, there are two issues when it comes to light-weighting models: (i) \textbf{Lack of global information extraction.} The inherent locality of convolutional operations limits the capture of global context. Although recent methods have used large-kernel convolutions to extract global information~\cite{tang2024cmunext, lee20223d}, they fundamentally lack an understanding of broader context relationships and suffer from the problem of feature redundancy~\cite{si2024scsa}. Hybrid CNN-Transformer architectures~\cite{chen2021transunet, qi2022medt, cao2022swin} have partially addressed these issues, but they require a large amount of medical data and computational resources, and their complex designs impede practical deployment and lightweight applications. \textbf{(ii)  Fail to fully consider the issues of channel redundancy in medical imaging and the variation in lesion scales}. Channel redundancy refers to the phenomenon that there is repeated or overlapping information in the feature representation of different channels in the convolutional layer~\cite{zhang2020split}. As shown in Fig.~\ref{fig:svd}, even within a single dataset, we observe significant differences in the size of the foreground (Fig.~\ref{fig:svd} (a)). Meanwhile, classical models also exhibit channel redundancy (Fig.~\ref{fig:svd} (c)). We've plotted the density distribution of the ratio of foreground lesion-pixel count to total image-pixel count for different datasets in Fig.~\ref{fig:svd} (a). There is a considerable portion of data where the foreground is close to or even exceeds the scale of the background, which will pose more difficulties for models lacking global scale information. We used the same SVD algorithm to process channel dimension and presented the average number of channel singular values of different models in Fig.~\ref{fig:svd} (c), which demonstrates high channel redundancy of previous models. 

\begin{figure}[htbp]
\centering
\includegraphics[width=0.45\textwidth]{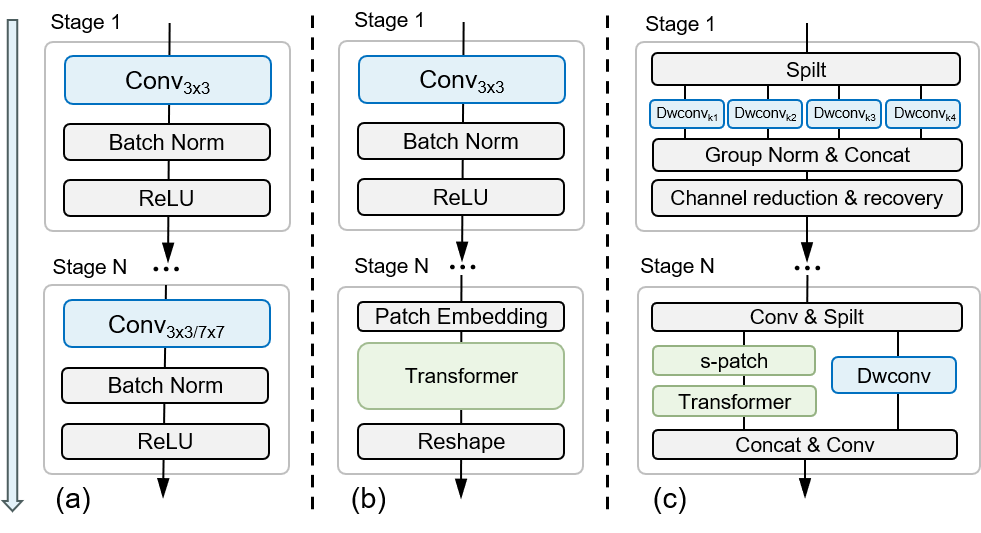} 
\caption{Comparison of our model with other mainstream methods. We simplify the details to highlight the differences. (a) A typical U-Net structure usually uses uniformly sized convolution kernels or larger convolution kernels at the bottom. (b) U-Net structure with the bottom replaced by VIT-related modules. (c) Our dual multiscale framework includes local multiscale multi-kernel convolution layers and global multiscale mixing layers.}
\label{fig:model_comparsion}
\end{figure}

To address these challenges, we propose LGMSNet (Fig.~\ref{fig: LGMSNet}), a simple yet effective lightweight framework for universal medical segmentation through local multiscale and global multiscale. Fig.~\ref{fig:model_comparsion} illustrates the differences in the main architectures between our model and the classical models.
For local multiscale processing, we design a deep convolutional module with heterogeneous kernel sizes to extract high-frequency differential features between foreground and background, enhancing sensitivity to lesions of varying scales while mitigating channel redundancy (as shown in Fig.~\ref{fig:svd} (c)). For global multiscale processing, we introduce a sparse hybrid Transformer-conv branch to capture deep low-frequency information, thereby achieving global information extraction of feature maps with local inductive bias while avoiding parameter explosion. Comparative analyses confirm our model's superior segmentation performance (Fig.~\ref{fig:svd} (b)) with reduced redundancy (Fig.~\ref{fig:svd} (c)) and enhanced foreground capture capability (Fig.~\ref{fig:svd} (d)). LGMSNet achieves modality-agnostic segmentation for both 2D and 3D data, validated through extensive experiments across four 2D datasets and one 3D dataset. Results demonstrate LGMSNet's exceptional segmentation accuracy, parameter efficiency, and cross-domain generalization.

Our key contributions include: 
\begin{itemize}
\item  \textbf{Local-level multiscale:} An intra-layer multiscale convolutional module that adaptively captures multiscale foreground features while reducing channel redundancy. 
\item \textbf{Global-level multiscale:} A sparse hybrid Transformer-convolution branch that efficiently learns global contexts with convolutional inductive biases. 
\item  A lightweight universal architecture that demonstrates state-of-the-art performance and generalization across both 2D and 3D modalities with minimal parameters.
\end{itemize}

\section{Related work}
\subsection{CNN and Transformer-based Segmentation Work}

With the development of deep learning, the encoding and decoding convolutional network represented by U-net~\cite{ronneberger2015u} has first replaced traditional machine learning algorithms~\cite{tsai2003shape}. The U-Net network has become the mainstream method for medical image segmentation with its simple structure and excellent image segmentation performance and a large number of related variants have thus emerged ~\cite{zhou2019unet++, huang2020unet, cciccek20163d}. 
3D-Unet~\cite{cciccek20163d} extends the U-net segmentation method by using 3D convolution to achieve the segmentation of 3D data. Although CNN-based methods have made significant progress in medical image segmentation, the inherent limitations of convolutional operations (such as lack of global information and existence of feature redundancy~\cite{si2024scsa}) hinder the further improvement of segmentation performance.

\begin{figure*}[htbp]
\centering
\includegraphics[width=0.95\textwidth]{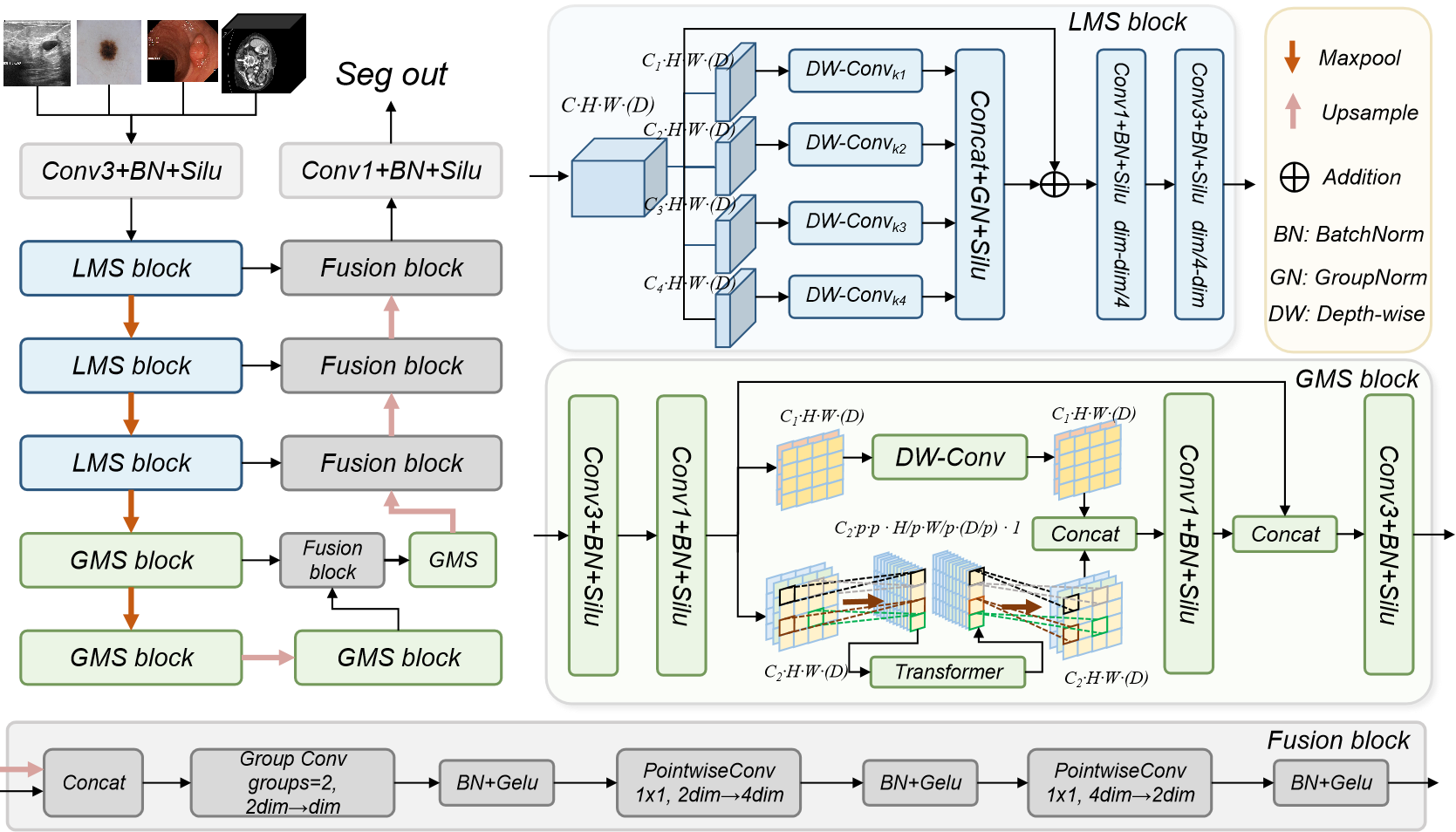}
\caption{The LGMSNet framework. It mainly consists of a Local multiscale (LMS) block that utilizes channel redundancy to extract high-frequency scale differences and a Global multiscale (GMS) block that perceives low-frequency contextual information with local inductive bias.} 
\label{fig: LGMSNet}
\end{figure*}

With the success of Transformers in the field of computer vision~\cite{dosovitskiy2020image}, more Transformer-based methods have emerged for medical image segmentation.
TransUNet~\cite{chen2021transunet} combines CNNs and Transformers to increase the global awareness of convolutional networks. UNETR~\cite{hatamizadeh2022unetr} uses a transformer as an encoder to learn the sequence representation of the input volume and follows the U-Net structure to achieve 3D segmentation. Compared with UNETR, SwinUNETR~\cite{hatamizadeh2021swin} and SwinAttUNet~\cite{li2023new}  use Swin transformer to achieve multi-resolution attention computing interaction. However, Transformer-based networks often require a large amount of computation, do not fully exploit the inherent correlation between features~\cite{li2024dual}, and their lack of local inductive bias hinders the perception of detailed features.

\subsection{Lightweight segmentation work}
Lightweight medical image segmentation has attracted a lot of attention due to its applicability in resource-constrained environments~\cite{chen2024tinyu}. UNeXt~\cite{valanarasu2022unext} uses depth-separable convolution reduction parameters for lightweight segmentation. ConvUNeXt~\cite{han2022convunext} further integrates lightweight attention to enhance segmentation performance. Recently, methods that use large-kernel convolution as a substitute for the Transformer to extract global context information have appeared in more and more lightweight networks. ConvMixer~\cite{trockman2022patches} and RepLKNet~\cite{ding2022scaling} use large convolutional kernels to process long-range spatial location information, but these methods are designed for natural images. CMUNeXt~\cite{tang2024cmunext} uses large kernels and inverted bottlenecks to fuse long-range context information for medical images. Although existing lightweight networks offer a promising approach to addressing the global healthcare gap, due to the reduction in the number of parameters and the decrease in feature representation ability, they often struggle to outperform the current state-of-the-art models and have reduced generalization capabilities~\cite {chen2024tinyu}.


\definecolor{lightgreen}{RGB}{144, 238, 144} 
\definecolor{green}{RGB}{189, 230, 205}
\definecolor{blue}{RGB}{228,238,188}          
\definecolor{gray}{RGB}{255,248,197} 

\begin{table*}[htbp]
\caption{Results on 2D dataset. The numbers without subscripts are from CMUNeXt \cite{tang2024cmunext} and the subscripts show standard deviations between replicate experiments. \colorbox{green}{Color} represents the top 1 result, \colorbox       {blue}{color} represents the top 2, and \colorbox{gray}{color} represents the top 3 (heavyweight models and lightweight models are ranked separately) and the unit of param is one million. The following table settings are the same.}  
\label{tab:result1}  
\resizebox{1\linewidth}{!}
{
\begin{tabular}{l|lll|ll|ll|cc|cc}
\hline
\multirow{3}{*}{Networks} & \multirow{3}{*}{Param} & \multirow{3}{*}{FPS} & \multirow{3}{*}{Gflops} & \multicolumn{4}{c|}{Ultrasound}                                  & \multicolumn{2}{c|}{Dermoscopy} & \multicolumn{2}{c}{Colonoscopy} \\ \cline{5-12} 
                          &                         &                         &                      & \multicolumn{2}{c|}{BUSI}          & \multicolumn{2}{c|}{TNSCUI} & \multicolumn{2}{c|}{ISIC18}     & \multicolumn{2}{c}{Kvasir}      \\ \cline{5-12} 
                          &                         &                         &                      & IoU(\%)   & \multicolumn{1}{c|}{F1(\%)}    & IoU(\%)          & F1(\%)           & IoU(\%)             & F1(\%)            & IoU(\%)             & F1(\%)            \\  \hline
U-Net\cite{ronneberger2015u}           & 34.52M  & 139.3  & 65.52 & 68.61                                   & 76.97                                     & 75.88                                     & 84.24                                     & 82.18$_{0.9}$                           & 89.91$_{0.5}$                           & 86.52$_{1.2}$                           & 91.90$_{0.9}$ \\
Atten U-Net\cite{oktay2018attention}     & 34.87M  &129.9  & 66.63 & 68.55                                   & 76.88                                     & 75.83                                     & 84.28                                     & 82.36$_{0.2}$                           & 89.24$_{0.1}$                           & \cellcolor{gray}87.14$_{0.2}$           & \cellcolor{gray}92.44$_{0.2}$ \\
U-Net++\cite{zhou2019unet++}         & 26.90M  & 125.5  &37.62  & 69.49                                   & 78.06                                     & 76.90                                     & 85.13                                     & 82.29$_{0.1}$                           & 89.17$_{0.1}$                           & 86.60$_{0.4}$                           & 92.08$_{0.4}$ \\
Swin-Unet\cite{cao2022swin}       & 27.14M  & 392.2  & 5.91  & 63.59                                   & 76.94                                     & 75.77                                     & \cellcolor{gray}85.82                     & 82.15$_{1.4}$                           & \cellcolor{gray}89.98$_{0.9}$           & 73.58$_{1.2}$                           & 82.17$_{0.9}$ \\
TransUnet\cite{chen2021transunet}       & 105.3M  & 112.9  & 38.52 & \cellcolor{gray}71.39                                   & \cellcolor{gray}79.85                                     & \cellcolor{gray}77.63                     & 85.76                                     & 83.17$_{1.3}$           &\cellcolor{green}\textbf{90.57$_{0.7}$} & 85.07$_{0.5}$                           & 90.62$_{0.4}$ \\
MissFormer\cite{huang2022missformer}      & 35.45M  & 151.4  & 7.25  & 63.29$_{3.0}$                           & 73.47$_{3.1}$                            & 68.26$_{2.4}$                             & 76.71$_{0.2}$                             & 80.99$_{0.3}$                           & 88.03$_{0.3}$                           & 86.37$_{0.6}$                           & 91.79$_{0.5}$ \\
UCTransNet\cite{wang2022uctransnet} & 66.24M &86.67 & 43.06 & 70.05$_{2.3}$ & 78.51$_{2.5}$ &75.51$_{0.3}$ & 84.08$_{0.2}$  & 82.76$_{0.4}$ & 89.50$_{0.3}$ &\cellcolor{green}\textbf{87.97$_{0.4}$} & \cellcolor{green}\textbf{92.93$_{0.2}$}  \\
nnUNet\cite{isensee2021nnu}          & 26.10M  & ---    & 12.67 & \cellcolor{blue}72.11$_{3.5}$           & \cellcolor{blue}80.09$_{3.7}$             & \cellcolor{blue}78.99$_{0.1}$   & \cellcolor{blue}86.85$_{0.2}$   & \cellcolor{blue}83.31$_{0.6}$           & 89.84$_{0.5}$                           & 85.60$_{0.1}$   &91.50$_{0.1}$  \\
MADGnet\cite{nam2024modality}        & 31.56M    &  35.53      &  10.08      &  \cellcolor{green}\textbf{72.92$_{2.6}$}                           &  \cellcolor{green}\textbf{81.04$_{2.7}$}                             &  \cellcolor{green}\textbf{79.08$_{0.1}$}                             &   \cellcolor{green}\textbf{86.92$_{0.1}$}   &    \cellcolor{green}\textbf{83.72$_{0.2}$}                          &  \cellcolor{blue}90.21$_{0.1}$    &  \cellcolor{blue}87.61$_{0.1}$                           & \cellcolor{blue}92.71$_{0.1}$ \\ \hline

MobileViT-s\cite{mehta2021mobilevit}     & 10.66M  & 301.1  & 2.13  & 64.28$_{3.8}$                           & 74.68$_{3.8}$                             & 71.64$_{0.2}$                             & 81.60$_{0.3}$                             & 80.12$_{0.4}$                           & 87.89$_{0.4}$                           &  84.21$_{0.1}$	        &  90.72$_{0.7}$\\
CMUNeXt\cite{tang2024cmunext}         & 3.14  M   &  471.4   &7.41 &\cellcolor{blue}71.56                   & \cellcolor{blue}79.86                     & \cellcolor{blue}77.01                                     & \cellcolor{blue}85.21                                     & \cellcolor{blue}82.49$_{0.3}$                           & 89.36$_{0.2}$                           & \cellcolor{blue}87.92$_{0.3}$           & \cellcolor{blue}92.94$_{0.4}$ \\
MedT\cite{qi2022medt}            & 1.37 M   & 22.97  & 2.40  & 63.36                                   & 73.37                                     & 71.00                                     & 80.87                                     & 81.79$_{0.9}$                           &  \cellcolor{gray}89.74$_{0.5}$                           & 79.70$_{1.2}$                           & 87.40$_{0.9}$ \\
UniRepLKNet\cite{ding2024unireplknet}     & 5.83  M   & 178.3  & 9.39  & 65.26$_{1.7}$                           & 73.96$_{1.9}$                             & 67.73$_{0.9}$                             & 77.20$_{0.9}$                             & 81.64$_{0.2}$                           & 88.82$_{0.2}$                           & 85.30$_{1.0}$                           & 91.18$_{0.8}$ \\
RepViT-m3\cite{wang2024repvit}       & 14.37M  & 238.9  & 2.79  & 56.07$_{3.5}$                           & 67.62$_{3.9}$                             & 66.21$_{0.6}$                             & 77.09$_{0.5}$                             & 78.34$_{0.6}$                           & 86.62$_{0.5}$                           & 81.55$_{1.2}$                           & 88.75$_{0.9}$ \\

UNeXt\cite{valanarasu2022unext}           & 1.47  M   & 650.4  & 0.58  & 65.04$_{2.7}$                           & 74.16$_{2.8}$                             & 71.04$_{0.1}$                             & 80.46$_{0.2}$                             &  \cellcolor{gray}82.10$_{0.9}$                           & \cellcolor{blue}89.93$_{0.5}$                           & 84.78$_{0.8}$                           & 91.01$_{0.7}$ \\

TinyUnet\cite{chen2024tinyu}        & 0.48 M       &359.5        & 1.66      & 66.21$_{2.7}$                           & 75.01$_{2.6}$                             & 74.03$_{0.2}$                             & 82.95$_{0.1}$                             & 81.95$_{0.3}$                           & 88.99$_{0.2}$                           &  \cellcolor{gray}85.47$_{0.1}$                           &  \cellcolor{gray}91.35$_{0.1}$ \\
ERDUnet\cite{li2023erdunet}  & 10.21M    & 137.6    & 10.29   & \cellcolor{gray}68.67$_{1.9}$  &   \cellcolor{gray}77.74$_{2.0}$ &  \cellcolor{gray}76.02$_{0.4}$      &  \cellcolor{gray}84.63$_{0.3}$                           & 82.03$_{0.2}$                            &88.96$_{0.2}$                            &    84.93$_{0.5}$                         & 90.88$_{0.6}$  \\
LGMSNet(ours)            & 2.32 M       & 302.8       & 4.89      & \cellcolor{green}\textbf{73.40$_{2.0}$} & \cellcolor{green}\textbf{81.82$_{1.9}$}   & \cellcolor{green}\textbf{78.03$_{0.3}$}             & \cellcolor{green}\textbf{86.17$_{0.3}$}             & \cellcolor{green}\textbf{83.46$_{0.3}$} & \cellcolor{green}\textbf{89.93$_{0.2}$}           & \cellcolor{green}\textbf{88.14$_{0.2}$} & \cellcolor{green}\textbf{93.04$_{0.3}$} \\
\hline
\end{tabular}
}

\end{table*}

\section{Method}
Fig.~\ref{fig: LGMSNet} illustrates the details of our proposed LGMSNet and its corresponding key modules. We first introduce the LMS block (see Sec.~\ref{subsec: LMS}), a convolutional module that leverages channel redundancy information to obtain co-level multiscale information. Next, we present the GMS block (see Sec.~\ref{subsec: GMS}), a combination of convolution and Transformer designed to capture global contextual information with a local inductive bias. Finally, we introduce LGMSNet (see Sec.~\ref{subsec: LGMSNet}), a lightweight model capable of performing universal 2D and 3D medical image segmentation.
\subsection{LMS Block} %
\label{subsec: LMS}
To address channel redundancy in medical imaging~\cite{lee20223d} and leverage the scale variation of foreground targets across intra-/inter-domain modalities, we propose a lightweight yet effective local multiscale feature extraction strategy. As illustrated in Fig. \ref{fig: LGMSNet}, given an input tensor \( X \in \mathbb{R}^{C \times H \times W} \), we first decompose it along the channel dimension into four branches through:

\begin{equation}
    X_i = \text{Split}(X) \in \mathbb{R}^{C_i \times H \times W},
\end{equation}
where \( i = 1,2,3,4 \) and \( \sum_{i=1}^4 C_i = C  \).

Each partition \( X_i \) undergoes deep convolution with a kernel size of \( k_i \) to capture semantic features at different scales while maintaining parameter efficiency. Compared to the single-size kernel convolution, our deep multi-kernel convolution can learn local inductive biases of different foreground scales more flexibly. After obtaining multiscale features, we stitch the output along the channel dimensions. To train stability and preserve different scale features, we use group normalization. And then we add the SiLU activation function, and set the output residual learning for enhancement:

\begin{equation}
    X'_i = \text{DWConv}_{k_i}(X_i),    
\end{equation}
\begin{equation}
    X'_{concat} = \text{Concat}(X'_1, X'_2, X'_3, X'_4),
\end{equation}
\begin{equation}
    X' = \text{SILU}(\text{GN}(X'_{concat})) + X,
\end{equation}
In order to dynamically recalibrate channel importance and suppress redundant background signals, we adopt a 1×1 convolution and reduce the number of channels to one-fourth of the input, and then combine BatchNorm and SiLU activation to ensure stable optimization and enhance nonlinear representation. The subsequent 3×3 convolution further promotes the fusion of cross-scale feature integration and local convolution bias, and increases the channel dimension to the number of input channels. The proposed LMS block enables efficient multiscale modeling, has low architectural complexity, and can be seamlessly integrated into various network architectures.

\begin{figure}[htbp]
\centering
\includegraphics[width=0.5\textwidth]{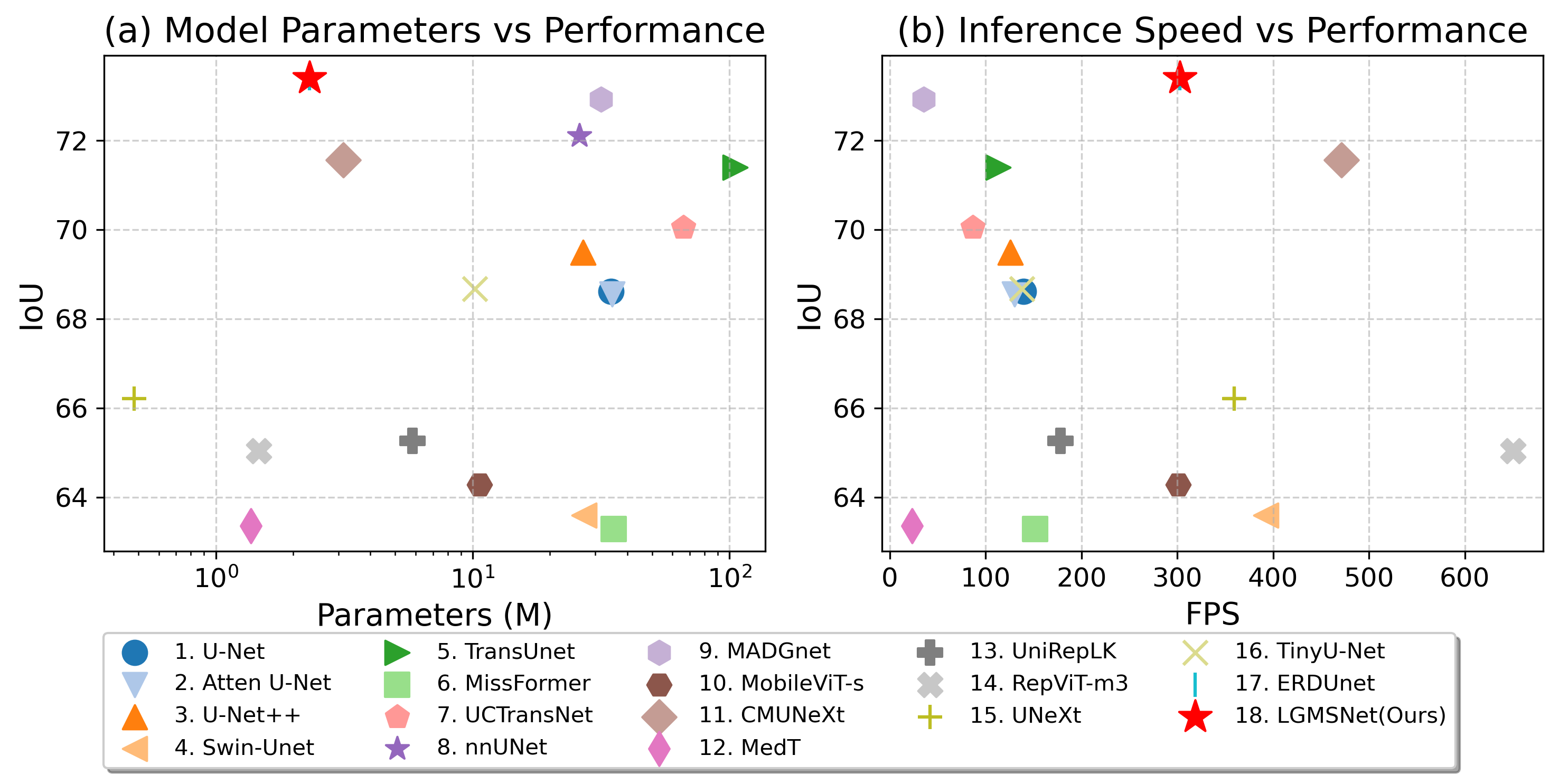}
\caption{Visual comparison results of parameter quantity, inference speed, and computational complexity on the BUSI dataset.}
\label{fig:performance}
\end{figure}

\subsection{GMS Block}
\label{subsec: GMS}
The inductive bias of CNNs~\cite{ronneberger2015u} and the long-range representational capabilities of Transformers~\cite{dosovitskiy2020image} play a crucial role in achieving superior performance in visual tasks~\cite{tang2024hyspark}. Still, convolution is often ignored in VIT-type architectures, while the latter is difficult to apply in lightweight networks due to its computational complexity. In order to take advantage of both in a lightweight network, inspired by MobileViT~\cite{mehta2021mobilevit}, we propose the sparse hybrid Transformer-Conv block from a global multiscale perspective, as illustrated in Fig. \ref{fig: LGMSNet}. 

Given an input tensor \( X \in \mathbb{R}^{C \times H \times W} \), the GMS block first encodes local spatial patterns through a 3×3 convolution, followed by channel expansion via 1×1 convolution to obtain enriched features \( X^E \). The expanded features are partitioned along channels into dual pathways: \( X^C \) for convolutional processing and \( X^T \) for transformer-based modeling.
\begin{equation}
    X^C\in\mathbb{R}^{C_1 \times H \times W}, X^T\in\mathbb{R}^{C_2 \times H \times W} = \text{Spilt}(X^E)
\end{equation}

This design allows the module to leverage convolutions' inductive bias and Transformers' global representational power. In the convolutional branch, we directly apply depthwise convolution to \( X^C \). In the Transformer branch, each channel of the input \( X^T \) is divided into patches \((h = w = p)\). The features of the corresponding positions of each block in the same channel are connected to obtain \( X^{T'} \):
\begin{equation}
    X^{T'} \in \mathbb{R}^{C_2 \cdot p \cdot p \times H/p \cdot W/p \times 1}=\text{Reshape}(X^T),
\end{equation}
Through this sparse method, the parameters and computational consumption are reduced. Since \( X^T \) is a local convolutional representation of \( X^E \), each pixel in the output of the Transformer branch encodes the information of all pixels in \( X^E \), helping the model capture the global information of each feature map and the attention relationship of different channels. Finally, the outputs \( X^T \) and \( X^C \) of the two branches are connected and fused using a 1x1 convolutional layer. The input \( X\) and output \( X' \) are then merged and processed with a 3x3 convolutional layer to obtain the final output.

\begin{equation}
    X'   = \text{Concat}\left(\text{DWConv}(X^C), \text{Transformer}(X^T)\right),
\end{equation}
\begin{equation}
    X_{\text{out}} = \text{Conv}_3\left(\text{Concat}\left(\text{Conv}_1(X'), X\right)\right),
\end{equation}

\subsection{LGMSNet}
\label{subsec: LGMSNet}

As illustrated in Fig. \ref{fig: LGMSNet}, our proposed LGMSNet model is based on the U-Net encoder-decoder architecture, utilizing MaxPooling for four downsampling operations and linear interpolation to restore resolution. To balance efficiency and performance, we employ a convolutional structure based on the LMS block at high resolutions, leveraging redundant channel information while extracting multiscale information from different data. At lower resolutions, we use the hybrid GMS block, combining the inductive bias of convolutions with the global representational capabilities of Transformers to provide channel information interaction and global spatial awareness to the network. Like CMUNeXt, we incorporate a Fusion block that combines group convolutions and pointwise convolutions to merge corresponding features from the encoder and decoder.


\section{Experiments}

\begin{table*}[t]
\centering
\caption{Generalization Experiment of lightweight model on the Unseen Dataset (BUSI$\to$BUS, TNSCUI$\to$TUCC, ISIC18$\to$PH2, Kvasir$\to$CVC-300). The results are the average of repeated experiments. We also show independent sample t-test analysis ($\alpha$=0.05, one-tailed test) of our model and other models, where * indicates significant, ** indicates significant with p<0.01.} 
\label{tab:unseen}
\resizebox{0.9\textwidth}{!}{

\begin{tabular}{l|c|ll|ll|ll|ll}
\hline
\multirow{3}{*}{Networks} & \multirow{3}{*}{Volume} & \multicolumn{4}{c|}{Ultrasound} & \multicolumn{2}{c|}{Dermoscopy} & \multicolumn{2}{c}{Colonoscopy} \\
\cline{3 - 10} 
 &  & \multicolumn{2}{c|}{BUS} & \multicolumn{2}{c|}{TUCC} & \multicolumn{2}{c|}{PH2} & \multicolumn{2}{c}{CVC-300} \\
\cline{3 - 10} 
 &  &  \multicolumn{1}{c}{IoU(\%)} & \multicolumn{1}{c|}{F1(\%)} & \multicolumn{1}{c}{IoU(\%)} & \multicolumn{1}{c|}{F1(\%)}& \multicolumn{1}{c}{IoU(\%)} & \multicolumn{1}{c|}{F1(\%)} & \multicolumn{1}{c}{IoU(\%)} & \multicolumn{1}{c}{F1(\%)} \\ \hline

UNeXt                                         & MICCAI'22    & 69.02$^{**}$  & 79.27$^{**}$ &54.83$^{**}$&66.04$^{**}$ & 82.00$^{*}$  & 89.57$^{*}$ & 65.75$^{*}$  & 75.23$^{*}$ \\ 
ERDUnet	&TCSVT'24	         &\cellcolor{gray}71.75$^{**}$	&80.72$^{**}$&\cellcolor{gray}59.66$^{*}$&\cellcolor{gray}70.06$^{**}$	&82.90$^{*}$	&90.06$^{*}$	&72.23	&80.62 \\
RepViT-m3  & CVPR'24  & 65.07$^{**}$  & 76.38$^{*}$&57.28$^{**}$&68.88$^{*}$ & 79.95$^{**}$  & 88.27$^{**}$ &62.96$^{*}$  & 73.68$^{*}$ \\ 
MobileViT-s &ICLR'22 &70.70$^{**}$  &\cellcolor{gray}81.14$^{**}$& \cellcolor{blue}59.83$^{*}$& \cellcolor{blue}70.78$^{*}$ &83.19$^{*}$  &90.42$^{*}$  &\cellcolor{gray}72.50  &\cellcolor{gray}81.92 \\
UniRepLK & CVPR'24 & 66.19$^{*}$  & 75.65$^{*}$&55.53$^{**}$& 67.12$^{**}$ & \cellcolor{gray}83.83$^{*}$  & \cellcolor{blue}90.8  & 68.31$^{*}$  & 75.59$^{*}$     \\ 
CMUNeXt& ISBI'24  & \cellcolor{blue}74.49$^{*}$  & \cellcolor{blue}83.04$^{*}$&59.08$^{**}$&69.49$^{**}$ & \cellcolor{blue}83.85$^{*}$  & \cellcolor{gray}90.75 & \cellcolor{blue}74.57  & \cellcolor{green}\textbf{82.60} \\ 
TinyU-net  & MICCAI'24    & 64.49$^{*}$  & 73.7$^{*}$ &57.76$^{**}$&69.03$^{*}$ & 81.96$^{*}$  & 89.53$^{*}$ & 71.22  & 78.50  \\ 
\hline
LGMSNet     &              Ours            & \cellcolor{green}\textbf{76.86}  & \cellcolor{green}\textbf{85.36} &\cellcolor{green}\textbf{61.14}&\cellcolor{green}\textbf{71.73} & \cellcolor{green}\textbf{85.75}  & \cellcolor{green}\textbf{91.92} & \cellcolor{green}\textbf{74.73}  & \cellcolor{blue}82.45 \\ 
\hline
\end{tabular}
}
\end{table*}

\begin{figure*}[htbp]
\centering
\includegraphics[width=1\textwidth]{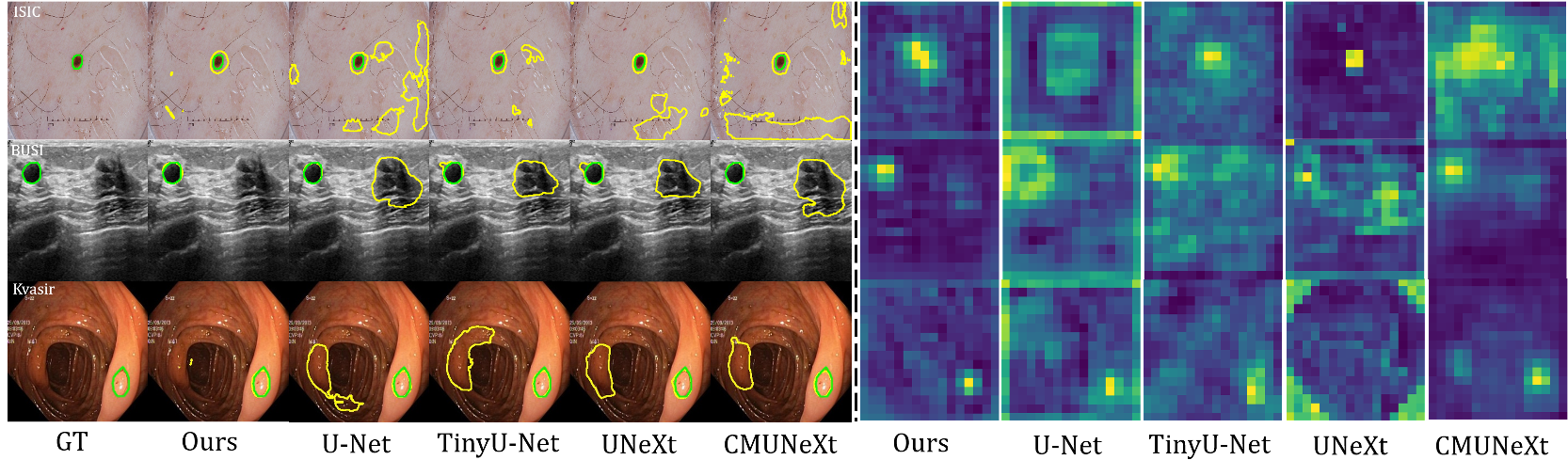}
\caption{2D data visualization results. The segmentation results show serious mis-segmentation of other models, and the feature map (four times downsampling) shows that our model has a stronger ability to focus on the lesion. \textcolor{green}{Gellow line} is ground truth, and the \textcolor{yellow}{yellow line} is the prediction.}
\label{fig:segout1}
\end{figure*}

\subsection{Datasets}

\textbf{BUSI dataset}: The Breast Ultrasound Images (BUSI) dataset~\cite{al2020dataset}, which was collected from 600 female patients in 2018, contains a total of 780 breast ultrasound images. The dataset covers 133 normal cases, 487 benign cases, and 210 malignant cases, each with corresponding ground truth labels. Following the approaches of recent studies~\cite{valanarasu2022unext,al2020dataset}, we only utilized the benign and malignant cases from this dataset.
\\ \textbf{TNSCUI dataset}: The Thyroid Nodule Segmentation and Classification in Ultrasound Images 2020 (TNSCUI) dataset, available at https://tn-scui2020.grand-challenge.org/Home/, was collected by the Chinese Artificial Intelligence Alliance for Thyroid and Breast Ultrasound (CAAU). It includes 3644 cases from different ages and genders. The dataset comprises both benign and malignant cases, each with corresponding ground truth labels.
\\ \textbf{ISIC dataset}: The ISIC 2018 dataset~\cite{gutman2016skin} was published by the International Skin Imaging Collaboration (ISIC) as a large-scale dataset of dermoscopy images. This Task 1 dataset is a challenge in lesion segmentation. It contains 2594 images of skin lesions.
\\ \textbf{Kvasir dataset}: The Kvasir-SEG dataset~\cite{jha2020kvasir} contains 1000 polyp images and their corresponding ground truth from the Kvasir Dataset.
\\ \textbf{BTCV dataset}: The Beyond The Cranial Vault (BTCV) dataset~\cite{landman2015miccai} is an essential 3D abdominal organ segmentation benchmark dataset, covering 13 organs, and the annotations are completed by professional radiologists.
\\ \textbf{KiTS23 dataset}: The Kidney and Kidney Tumor Segmentation 2023 (KiTS23) dataset ~\cite{heller2023kits21} consists of 599 cases, focusing on segmenting three categories: the kidney, tumors, and cysts, and serves as an important benchmark for advancing kidney and tumor segmentation techniques. 

Unlike most previous works, in order to test the cross-dataset robustness of the model, we conducted a generalization experiment on unseen validation data including BUS~\cite{zhang2022busis} (562 breast ultrasound images), TUCC~\cite{tucc} (17K images of thyroid nodule patients), PH2~\cite{mendoncca2013ph} (200 dermoscopic images), and CVC-300~\cite{vazquez2017benchmark} (912 colonoscopy images).

\begin{table*}[htbp]
\centering
\caption{Dice results of BTCV abdominal dissection. Our model achieves optimal performance with minimal parameters and has greatly improved gall segmentation.}
\label{tab:btcv}
\resizebox{1\textwidth}{!}{

\begin{tabular}{c |cc | ccccccccccc | c}
\hline 
\multicolumn{1}{c |}{Method} &Volume & \multicolumn{1}{l |}{Param} & \multirow{1}{*}{Spl} & \multirow{1}{*}{RKid} & \multirow{1}{*}{Lkid} & \multirow{1}{*}{Gall} & \multirow{1}{*}{Eso} & \multirow{1}{*}{Liv} & \multirow{1}{*}{Sto} & \multirow{1}{*}{Aor} & \multirow{1}{*}{IVC} &Veins &Pan  & \multirow{1}{*}{Avg} \\
\hline 
SegMamba\cite{xing2024segmamba}  & MICCAI'24 & 65.18                & \cellcolor{green}\textbf{96.11}  & \cellcolor{green}\textbf{95.02}  &94.58  & 61.46 & \cellcolor{blue}75.26 & \cellcolor{green}\textbf{97.08} & \cellcolor{blue}85.49 & \cellcolor{gray}89.74 & \cellcolor{blue}86.00   & 75.88 & \cellcolor{green}\textbf{83.33}& \cellcolor{blue}83.19 \\ 

MedNeXt\cite{roy2023mednext}   & MICCAI'23 & \cellcolor{blue}10.51                  & 95.75  &  \cellcolor{blue}94.92  & \cellcolor{gray}94.60  & \cellcolor{gray}63.09 & \cellcolor{green}\textbf{77.87} &  \cellcolor{gray}96.92 & 77.48 & 89.64 & \cellcolor{gray}85.55  & \cellcolor{blue}76.31 & 82.64 &  \cellcolor{gray}82.71 \\ 
Swin UNETR\cite{hatamizadeh2021swin} &MICCAI'21& 61.98                 & \cellcolor{blue}96.07  & 94.83  &\cellcolor{green}\textbf{94.85} & 60.47 &  \cellcolor{gray}75.03 & \cellcolor{blue}96.99 & \cellcolor{gray}82.77 &  \cellcolor{green}\textbf{90.51} & 84.95  & \cellcolor{green}\textbf{77.43} & \cellcolor{gray}82.65 & 82.57 \\ 
UNETR\cite{hatamizadeh2022unetr}  &WACV'22   &92.61                  & 91.45  & 94.35  & 93.49  & 62.42 & 69.80 & 95.66 & 76.12 & 88.04 & 80.96  & 69.76 & 75.05  & 79.03 \\ 

3D UX-Net\cite{lee20223d}   & ICLR'23 &\cellcolor{gray}53.01            & 95.20  &	94.67  & 94.36  &  \cellcolor{blue}72.92 & 74.72 & 96.61 & 81.99 & 89.45 & 85.00 & 74.48 & 78.76   & 82.56 \\ 
LGMSNet      & Ours&\cellcolor{green}\textbf{9.86}    & \cellcolor{gray}95.82  & \cellcolor{gray}94.84  & \cellcolor{blue} 94.70  &\cellcolor{green}\textbf{ 79.37} & 73.38 & \cellcolor{blue}96.99 & \cellcolor{green}\textbf{85.80} & \cellcolor{blue}90.44 &\cellcolor{green}\textbf{86.69}  &\cellcolor{gray}76.15 	&\cellcolor{blue}83.06 	 
 & \cellcolor{green}\textbf{84.10} \\ 
\hline
\end{tabular}
}
\end{table*}

\begin{figure*}[htbp]
\centering
\includegraphics[width=1\textwidth]{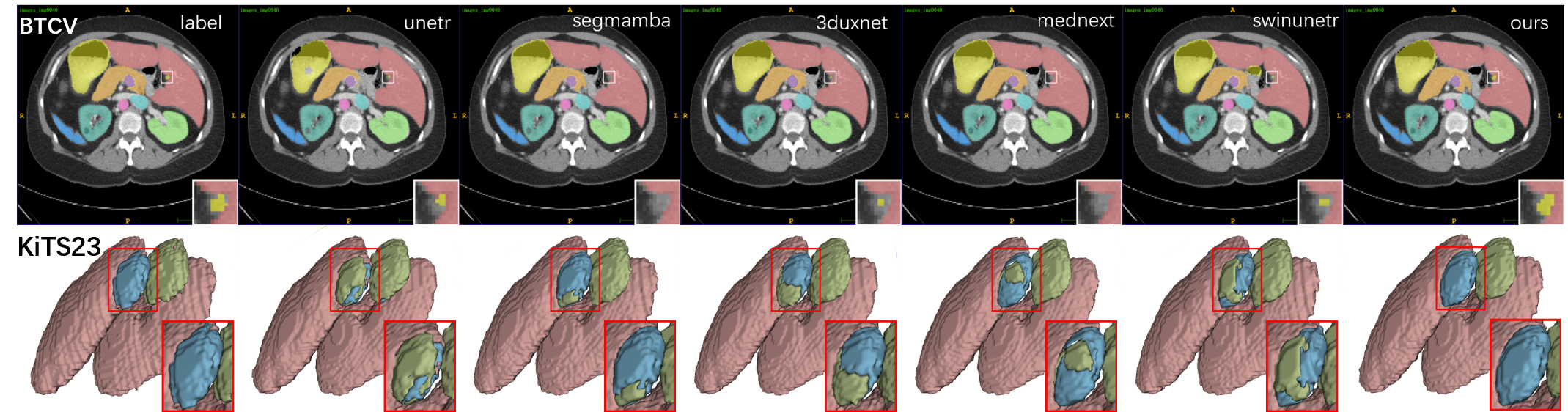}
\caption{3D data segmentation visualization results. The first row is the slice visualization result of BTCV, and the lower right corner shows the enlarged view of the bile organ. The second row is the 3D segmentation result of KiTS23, and the lower right corner shows the enlarged view of the Tumor.}
\label{fig:btcv_virual}
\end{figure*}
\subsection{Experimental Setup}

For the BUSI, TNSCUI, and ISIC datasets, we followed the three data splitting methods described in~\cite{tang2024cmunext} and reported the average results. For the Kvasir dataset, we adopted the splitting approach described in~\cite{articleducknet}, selecting the model based on the validation set and reporting the average results over three test sets. We implemented the LGMSNet using PyTorch and employed a classic loss function for segmentation experiments. Specifically, for the ground truth \( y \) and the predicted output \( \hat{y} \), the loss function is defined as:
\begin{equation}
     \mathcal{L}=0.5\times BCE(\widehat{y},y)+Dice(\widehat{y},y),
\end{equation}
where BCE denotes the binary cross-entropy loss and Dice denotes the Dice loss.

We resized all training cases of the four 2D datasets to \( 256\times256 \) and applied random rotation and flipping for data augmentation. The initial learning rate was set to 0.02, and the number of training epochs was 500. For 2D data, we evaluated the segmentation performance using the intersection over union (IoU) and F1 score, following the metrics in~\cite{tang2024cmunext,valanarasu2022unext}. For 3D data, we followed the training framework and Dice evaluation metric of 3D UX-Net~\cite{lee20223d}. Detailed training settings can be found in the supplementary material and code.

\begin{table}[htbp]
\centering
\caption{Dice results of the KiTS23 dataset Segmentation. Our LGMSnet model achieves the highest average performance, demonstrating superior segmentation capabilities.}
\label{tab:kidney_tumor_cyst}
\resizebox{0.49\textwidth}{!}{
\begin{tabular}{c |c|ccc | c}
\hline 
Method &Param& Kidney & Tumor & Cyst & Avg \\ \hline
SegMamba\cite{xing2024segmamba} &65.18& \cellcolor{gray}93.47 & \cellcolor{gray}73.94 & \cellcolor{blue}61.95 &  \cellcolor{blue}76.45 \\
MedNeXt\cite{roy2023mednext}  & \cellcolor{blue}10.51&93.10 & 71.33 & 58.20 & 74.21 \\
Swin UNETR\cite{hatamizadeh2021swin} & 61.98&\cellcolor{green}\textbf{93.70} & \cellcolor{blue}74.01 & 60.36 &  \cellcolor{gray}76.02 \\
UNETR\cite{hatamizadeh2022unetr} & 92.61&91.96 & 55.74 & 47.55 & 65.08 \\
3D UX-Net\cite{lee20223d} &\cellcolor{gray}53.01& 93.29 & 69.90 & \cellcolor{green}\textbf{62.40} & 75.20 \\
LGMSNet & \cellcolor{green}\textbf{9.86}&\cellcolor{blue} 93.59 & \cellcolor{green}\textbf{77.37} &\cellcolor{gray} 61.61 & \cellcolor{green}\textbf{77.52}\\ \hline
\end{tabular}
}
\end{table}

\subsection{Comparison with SOTA Methods}
\label{sub:Comparsion}

We compared LGMSNet with mainstream state-of-the-art (SOTA) segmentation models~\cite{ronneberger2015u,zhou2019unet++,chen2021transunet} as well as lightweight natural image segmentation models~\cite{wang2024repvit,ding2024unireplknet}, as shown in Table~\ref{tab:result1}. Our model achieved leading segmentation performance across multiple modalities (Ultrasound, Dermoscopy, Colonoscopy), with significantly fewer parameters and lower computational complexity. Compared with the latest heavyweight SOTA model MADGNet, our model achieved comparable or even better performance with less than 10\% of the parameters and nearly 10$\times$ FPS. Compared with other lightweight models, our model achieved comprehensive superiority. We believe that this is due to the fact that LGMSNet can focus on the foreground position more accurately at an earlier stage compared to other models, which facilitates the model to accurately segment the area near the foreground, as shown in Fig.~\ref{fig:segout1}. These results demonstrate the efficiency and effectiveness of LGMSNet. The parameter quantity and FPS visualization in Fig~\ref{fig:performance} can be more intuitively observed to show the advantages of our model in comprehensive performance and speed.

Generalization is a key indicator of the practicality of lightweight models. We further evaluated the performance of LGMSNet and other lightweight models on unseen datasets (BUSI→BUS, TNSCUI→TUCC, ISIC18→PH2, Kvasir→CVC-300), as shown in Table~\ref{tab:unseen}. Our model consistently achieved the best performance across all modalities, with statistically significant improvements over other lightweight models (marked with * or **). Compared with the second-best model, our model achieved an average improvement of 2.0\% in IOU and 1.3\% in F1 score across the four unseen datasets. This demonstrates that our multiscale strategy is reproducibly robust across different data modalities. We also compared the lightweight model from the perspectives of network strategy, experimental data, etc. The results are shown in Table.~\ref{tab:duigou}. Compared with other models, our model has a better network strategy, more comprehensive segmentation modalities, and more extensive experimental comparison.

\begin{table}[htbp]
\centering
\caption{This table shows the performance of different methods in various aspects (2D, 3D, Global: global information, MS: whether the network implements a multiscale strategy, General: whether generalization experiments were conducted; modality: number of modalities, Eval sets) to better understand their characteristics and capabilities. \checkmark indicates satisfaction; $*$ indicates natural image models, which are not considered for comparison.}
\label{tab:duigou}
\resizebox{0.48\textwidth}{!}{
\begin{tabular}{c|c|c|c|c|c|c|c}
\hline
Method     & 2D             & 3D            & Global              & MS             & General           & Modality & Eval sets \\
\hline
RepViT-m3  &  \checkmark  &       &        &       & *   & *           & * \\
UniRepLK   &  \checkmark  &       &        &  \checkmark  & *  & *           & * \\
MobileViT-s&  \checkmark  &       &  \checkmark  &        & *  & *           & * \\
UNeXt      &  \checkmark  &       &  \checkmark  &          &   & 2     & 2 \\
ERDUnet    &  \checkmark  &       &  \checkmark  &          &    & 4    & 7 \\
CMUNeXt    &  \checkmark  &       &        &  \checkmark   &  & 1        & 4 \\
TinyU-net  &  \checkmark  &       &        &  \checkmark  &   & 2        & 2 \\
LGMSNet    &  \checkmark  &  \checkmark &  \checkmark  &  \checkmark    &  \checkmark & 4  & 10 \\
\hline

\end{tabular}
}
\end{table}

\begin{table*}[htbp]
\caption{Migration experiment of LMS block and GMS block.}
\label{tab: ablation}
\centering
\resizebox{0.78\textwidth}{!}{
\begin{tabular}{c|c|ll|ll|ll}
\hline
\multirow{2}{*}{Model} & \multirow{2}{*}{Param} & \multicolumn{2}{c|}{BUSI} & \multicolumn{2}{c|}{ISIC18} & \multicolumn{2}{c}{Kvasir} \\ 
\cline{3-8}
 & & IoU(\%) & F1(\%) & IoU(\%) & F1(\%) & IoU(\%) & F1(\%)  \\ \hline
TinyU-Net & 0.48  & 66.21        & 75.01        & 81.95 & 88.99 & 85.47 & 91.35  \\
     +LMS & 0.69  & 68.95 +4.14\%& 77.35 +3.12\%& 82.57 +0.76\% & 89.32 +0.37\%& 86.77 +1.52\%& 92.18 +0.91\% \\
     +GMS & 0.71  & 69.98 +5.70\%&78.79 +5.04\% & 82.22 +0.33\% & 89.09 +0.11\%& 87.25 +2.08\% & 92.69 +1.47\% \\ \hline
U-Net     & 34.53 & 68.61        & 76.97        & 82.18         & 89.91 & 86.52& 91.90 \\
     +LMS & 17.33 & 71.95 +4.87\%& 80.19 +4.18\%& 82.79 +0.74\% & 89.51 -0.44\%& 87.17 +0.75\% & 92.32 +0.46\%\\
     +GMS & 19.18 & 72.48 +5.64\%& 80.77 +4.94\%& 82.87 +0.84\% & 89.56 -0.38\%& 88.56 +2.36\%& 93.29 +1.51\%\\
\hline
\end{tabular}
}
\end{table*}

We extended the LGMSNet to 3D image segmentation, conducted tests on the BTCV and KiTS23 datasets, and compared the results with mainstream SOTA models. The results are shown in Table \ref{tab:btcv} and Table \ref{tab:kidney_tumor_cyst}. Our model only uses 15\% of the parameters of the SOTA model Segmama but achieves an average improvement of 1.1\% and 1.4\% in segmentation performance. For the gallbladder organ in the BTCV dataset, the segmentation result of our model is improved by 29.1\% compared with that of Segmamba and by 8.8\% compared with that of the second-ranked model, demonstrating the accurate segmentation ability for small organs. For the Tumor part of the KiTS dataset, LGMSNet outperforms Segmamba and the second-ranked model by 4.6\% and 4.5\% respectively, and can clearly segment the Tumor and Cyst, indicating the model's accurate learning of features of different categories. LGMSNet exhibits excellent performance with extremely few resources on both 2D and 3D datasets, highlighting its applicability and generalization ability in different segmentation tasks.

\subsection{Visualization Experiment}
\begin{figure}[htbp]
\centering
\includegraphics[width=0.5\textwidth]{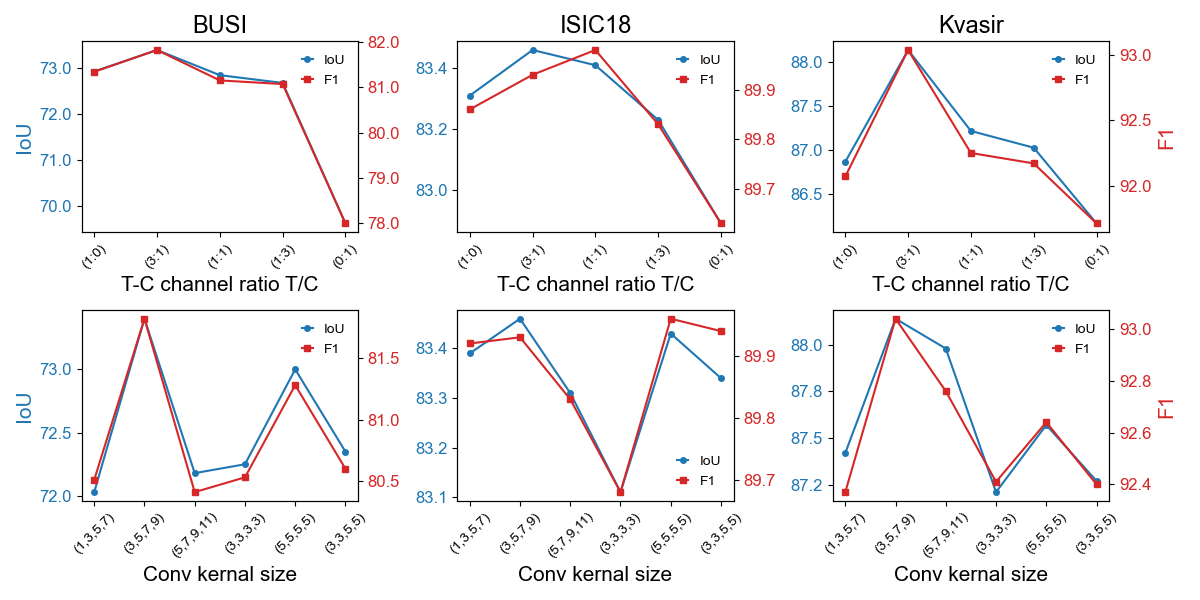}
\caption{Hyperparameter experiment of the model. We tested the convolution kernel groups of different sizes in the LMS block and the channel ratios processed by the Transformer-Conv in the GMS block. Comprehensively considering the performance of different datasets, the best effect is achieved when the T/C channel ratio is equal to 3:1 and the convolution combination is (3, 5, 7, 9).}
\label{fig:ablation}
\end{figure}

Fig.~\ref{fig:segout1} presents the visualization results of different models on the 2D datasets. The left side of Fig.~\ref{fig:segout1} compares the segmentation outcomes across three datasets, revealing significant missegmentation issues in other models. For the ISIC example, all models exhibited unexpected erroneous segmentation locations, yet our model demonstrated the smallest degree of error, followed by TinyUnet. In the BUSI example, all models except ours mistakenly identified the black shadow on the right side of the image as the lesion location, leading to incorrect segmentation. Similarly, for the Kvasir example, other models erroneously classified the protrusion on the left side of the image as a lesion. Our model performed well in all three scenarios, showcasing its ability to understand lesion details. We hypothesize that the misjudgments of other models are due to the lack of substantial global contextual information. The right side of Fig.~\ref{fig:segout1} displays the feature map results after four downsampling operations of the network. Compared with other models, our model exhibited a stronger ability to focus on the lesion foreground, indicating that it can detect lesion locations more precisely at an earlier stage.

Figure \ref{fig:btcv_virual} shows the segmentation visualization results of different models on two 3D datasets. In the slice visualization results of BTCV, the region of interest (ROI) of the gallbladder clearly demonstrates the differences among different models. SegMamba and MedNeXt can hardly segment the gallbladder region, while other models only identify part of the gallbladder region. For the 3D visualization results of KiTS, other models missegment some parts of the Tumor as Cyst, and only LGMSNet provides a clear segmentation. Overall, our model achieves the best performance, indicating that our multiscale strategy has better adaptability to organs of different sizes.

\subsection{Ablation Study}

To evaluate the effectiveness of the LMS and GMS modules and their hyperparameters, we conducted the following experiments:

1. LMS Module Convolutional Kernel Configuration Hyperparameters: We explored different convolutional kernel configurations within the LMS module. The experimental results are shown in Fig.~\ref{fig:ablation}. The combination of parameters (3, 5, 7, 9) yielded stable and optimal results, which, on average, outperformed traditional single-kernel convolution by 1\%.

2. GMS Module Transformer-Conv Branch Channel Ratio Hyperparameters: We considered five different ratio combinations, including pure Transformer and pure Conv settings. The results in Figure \ref{fig:ablation} indicate that the pure Conv setting performed the worst across all three datasets, highlighting the importance of incorporating global contextual information. However, the pure Transformer setting did not achieve the best performance either. Adding a certain proportion of convolutional channels to the model allows it to capture global features while retaining the ability to perceive local details.

3. Network Adaptability Analysis: We integrated the proposed LMS and GMS modules into the TinyU-Net and U-Net architectures by simply replacing the corresponding modules in the networks. For TinyU-Net, although the simple module replacement slightly increased the number of parameters, IOU and F1 score increased by 2.4\% and 1.8\% respectively. For U-Net, while reducing the number of model parameters by half, it can still improve the IOU and F1 score by 2.5\% and 1.7\% respectively, which demonstrates the transferability and generalization ability of the LMS and GMS modules.
\section{Conclusion}

LGMSNet achieves SOTA performance through synergistic local-global dual-level multiscale feature learning and addresses two critical limitations in existing methods - channel redundancy in convolutional operations and inefficient global context modeling. The proposed LMS block resolves scale variation challenges through heterogeneous kernel combination and channel-wise feature recalibration. In contrast, the GMS block enables computationally efficient global perception via sparse hybrid Transformer-convolution integration. Extensive validation across six public datasets demonstrates LGMSNet's superior segmentation accuracy. Notably, the model exhibits excellent cross-domain generalization capabilities on four unknown datasets, highlighting the potential of LGMSNet in promoting healthcare equity in the future.





\newpage
\bibliography{mybibfile}

\end{document}